\documentclass[a4paper,10pt,twocolumn]{article}
\usepackage[english]{babel}
\usepackage[latin1]{inputenc}
\usepackage{times}
\usepackage[T1]{fontenc}

\usepackage{graphicx}
\usepackage{amssymb}         
\usepackage{amsmath}
\usepackage{fancybox}
\usepackage{ragged2e}
\usepackage{colortbl}
\usepackage[gen]{eurosym} 
\usepackage{multirow}
\usepackage{textcomp} 
\usepackage{stmaryrd} 
\usepackage[a4paper]{geometry} 
\justifying

\newcommand{\intervalleff}[2]{\mathopen{[}#1\,;#2\mathclose{]}}

\newcommand{\iintervalleff}[2]{\mathopen{\llbracket}#1\,;#2\mathclose{\rrbracket}}

\renewcommand{\geq}{\geqslant}
\renewcommand{\leq}{\leqslant}

\usepackage{amssymb}
\usepackage{amsmath}

\usepackage{authblk} 

\usepackage[table]{xcolor}
\usepackage{soul,color}
\usepackage{multirow}
\usepackage[ruled,vlined, linesnumbered]{algorithm2e}
\SetKwRepeat{Do}{do}{while}
\usepackage{graphicx}
\usepackage{caption}
\usepackage{subcaption}
\usepackage{stmaryrd} 

\usepackage[latin1]{inputenc}
\usepackage[T1]{fontenc}

\definecolor{lightblue}{rgb}{0.8,0.9,1} 
\definecolor{lightred}{rgb}{1,0.5,0.4} 

\def\signed #1{{\leavevmode\unskip\nobreak\hfil\penalty50\hskip2em
  \hbox{}\nobreak\hfil#1%
  \parfillskip=0pt \finalhyphendemerits=0 \endgraf}}

\newsavebox\mybox

\usepackage[colorinlistoftodos]{todonotes}

\usepackage[normalem]{ulem}

\newcommand{\supp}[1]{}

\begin{document}
%
\title{Variations on Memetic Algorithms \\for Graph Coloring Problems}


\author[1]{Laurent Moalic\thanks{laurent.moalic@utbm.fr}}
\author[2]{Alexandre Gondran\thanks{alexandre.gondran@enac.fr}}
\affil[1]{{\small Univ. Bourgogne Franche-Comt\'e, UTBM, OPERA, Belfort, France}}
\affil[2]{{\small ENAC, French Civil Aviation University, Toulouse, France}}

\renewcommand\Authands{ and }

\date{}

\maketitle

\textbf{Abstract - }
Graph vertex coloring with a given number of colors is a well-known and much-studied NP-complete problem.
The most effective methods to solve this problem are proved to be hybrid algorithms such as 
memetic algorithms 
 or quantum annealing. 
Those hybrid algorithms use a powerful local search inside a population-based algorithm.
This paper presents a new memetic algorithm based on one of the most effective algorithms: the Hybrid Evolutionary Algorithm (\emph{HEA}) from Galinier and Hao (1999). 
The proposed algorithm, denoted \emph{HEAD} - for \emph{HEA} in Duet - works with a population of only two individuals.
Moreover, a new way of managing diversity is brought by \emph{HEAD}.
These two main differences greatly improve the results, both in terms of solution quality and computational time.
\emph{HEAD} has produced several good results for the popular DIMACS benchmark graphs,
  such as 222-colorings for <\texttt{dsjc1000.9}>, 81-colorings for <\texttt{flat1000\_76\_0}> and even 47-colorings for <\texttt{dsjc500.5}> and 82-colorings for <\texttt{dsjc1000.5}>.

\textbf{Keywords - } Combinatorial optimization, Metaheuristics, Coloring, Graph, Evolutionary 




\section{Introduction}
Given an undirected graph $G=(V,E)$ with $V$ a set of vertices and $E$ a set of edges,
 graph vertex coloring involves assigning each vertex with a color so 
that two adjacent vertices (linked by an edge) feature different colors.
The Graph Vertex Coloring Problem (GVCP) consists in finding the minimum number of colors, called \emph{chromatic number $\chi(G)$}, required to color the graph $G$ while respecting these binary constraints.
The GVCP is a well-documented and much-studied problem because this simple formalization can be applied to various issues such as 
frequency assignment problems~\cite{Aardal03, Dib10}, scheduling problems~\cite{Leighton79, Zufferey08, Wood69} 
and flight level allocation problems~\cite{Barnier04, Allignol2012}. 
Most problems that involve sharing a rare resource (colors) between different operators (vertices) 
can be modeled as a GVCP. 
The GVCP is NP-hard~\cite{Garey79}.

Given $k$ a positive integer corresponding to the number of colors,
a $k$-coloring of a given graph $G$ is a function $c$ that assigns to each vertex an integer between $1$ and $k$ as follows~:
$$
\begin{array}{rcl}
c:V&\to&\{1,2...,k\}\\
v&\mapsto&c(v)
\end{array}
$$
The value $c(v)$ is called the color of vertex $v$. 
The vertices assigned to the same color $i\in\{1,2...,k\}$ define a \emph{color class}, denoted $V_i$.
An equivalent view is to consider a $k$-coloring as a partition of $G$ into $k$ subsets of vertices: 
$c\equiv\{V_1,...,V_k\}$.

We recall some definitions~: 
\begin{itemize}
 \item a $k$-coloring is called \emph{legal} or \emph{proper $k$-coloring} if it respects the following binary constraints~: $\forall (u,v)\in E,\  c(u)\neq c(v)$.
Otherwise the $k$-coloring is called \emph{non legal} or \emph{non proper}; 
and edges $(u,v)\in E$ such as $c(u)= c(v)$ are called \emph{conflicting edges}, and $u$ and $v$ \emph{conflicting vertices}.  
 \item A $k$-coloring is a \emph{complete coloring} because a color is assigned 
to \emph{all} vertices; if some vertices can remain uncolored, the coloring is said to be \emph{partial}.
 \item An \emph{independent set} or a \emph{stable set} is a set of vertices, no two of which are adjacent.
It is possible to assign the same color to all the vertices of an independent set without producing any conflicting edge. 
The problem of finding a minimal graph partition of independent sets is then equivalent to the GVCP.
\end{itemize}

The $k$-coloring problem - finding a proper $k$-coloring of a given graph $G$ - is NP-complete~\cite{Karp72} for $k>2$.
The best performing exact algorithms are generally not able to find a proper $k$-coloring 
in reasonable time when the number of vertices is greater than $100$ for random graphs~\cite{Johnson91, Dubois93, Malaguti11}. Only for few bigger graphs, exact approaches can be applied successfully~\cite{Held2011}.
In the general case, for large graphs, one uses heuristics that partially explore 
the search-space to occasionally find a proper $k$-coloring in a reasonable time frame.
However, this partial search does not guarantee that a better solution does not exist. 
Heuristics find only an upper bound of $\chi(G)$ by successively solving the $k$-coloring problem with decreasing values of $k$.

This paper proposes two versions of a hybrid metaheuristic algorithm, denoted \emph{HEAD'} and \emph{HEAD}, integrating a tabu search procedure 
with an evolutionary algorithm for the $k$-coloring problem. 
This algorithm is built on the well-known Hybrid Evolutionary Algorithm (\emph{HEA}) of Galinier and Hao~\cite{Galinier99}.
However, \emph{HEAD} is characterized by two original aspects: the use of a population of only two individuals and an innovative way to manage the diversity.
This new simple approach of memetic algorithms provides excellent results on DIMACS benchmark graphs.


The organization of this paper is as follows. 
First, Section~\ref{sect:review} reviews related works and methods of the literature proposed for graph coloring 
and focuses on some heuristics reused in \emph{HEAD}.
Section~\ref{sect:HEAD} describes our memetic algorithm, \emph{HEAD} solving the graph $k$-coloring problem. 
The experimental results are presented in Section~\ref{sect:results}. 
Section~\ref{sect:analysis} analyzes why \emph{HEAD} obtains significantly better results than \emph{HEA} and 
investigates some of the impacts of diversification.
Finally, we consider the conclusions of this study and discuss possible future researches in Section~\ref{sect:conclusion}.

\section{Related works}\label{sect:review}
Comprehensive surveys on the GVCP can be found in~\cite{Galinier06, Galinier13, Malaguti10}. 
These first two studies classify heuristics according to the chosen search-space.
The Variable Space Search of~\cite{Hertz08} is innovative and didactic because it works with three different search-spaces.
Another more classical mean of classifying the different methods is to consider how these methods explore the search-space;
three types of heuristics are defined: constructive methods, local 
searches and population-based approaches.

We recall some important mechanisms of \emph{TabuCol} and \emph{HEA} algorithms because our algorithm \emph{HEAD} shares common features with these algorithms.
Moreover, we briefly present some aspects of the new Quantum Annealing algorithm for graph coloring denoted \emph{QA-col}~\cite{Titiloye11a,Titiloye11b,Titiloye12} which has 
 produced, since 2012, most of the best known colorings on DIMACS benchmark.

\subsection{TabuCol}
In 1987, Hertz and de Werra~\cite{Hertz87} presented the \emph{TabuCol} algorithm, one year after Fred Glover introduced the tabu search.
This algorithm, which solves $k$-coloring problems, was enhanced in 1999 by~\cite{Galinier99} and in 2008 by~\cite{Hertz08}. 
The three basic features of this local search algorithm are as follows:
\begin{itemize}
 \item \textit{Search-Space and Objective Function:} the algorithm is a \emph{$k$-fixed penalty strategy}.
This means that the number of colors is fixed and non-proper colorings are taken into account.
The aim is to find a coloring that minimizes the number of conflicting edges 
under the constraints of the number of given colors and of completed coloring 
(see \cite{Galinier13} for more details on the different strategies used in graph coloring).
 \item \textit{Neighborhood:} a $k$-coloring solution is a neighbor of another $k$-coloring solution 
if the color of only one conflicting vertex is different.
This move is called a critic 1-move. 
A 1-move is characterized by an integer couple $(v, c)$ where $v\in\{1,...,|V|\}$ is the vertex number and $c\in\{1,...,k\}$ the new color of $v$. 
Therefore the neighborhood size depends on the number of conflicting vertices.
 \item \textit{Move Strategy:} the move strategy is the standard tabu search strategy. 
Even if the objective function is worse, at each iteration, 
one of the best neighbors which are not inside the tabu list is chosen.
Note that all the neighborhood is explored. 
If there are several best moves, one chooses one of them at random. 
The tabu list is not the list of each already-visited solution because this is computationally expensive.
It is more efficient to place only the reverse moves inside the tabu list.
Indeed, the aim is to prevent returning to previous solutions, 
and it is possible to reach this goal by forbidding the reverse moves during a given number of iterations (i.e. the tabu tenure).
The tabu tenure is dynamic: it depends on the neighborhood size.
A basic aspiration criterion is also implemented: it accepts a tabu move to a $k$-coloring, 
which has a better objective function than the best $k$-coloring encountered so far.
\end{itemize} 
Data structures have a major impact on algorithm efficiency, 
constituting one of the main differences between the Hertz and de Werra version of 
\emph{TabuCol}~\cite{Hertz87} and the Galinier and Hao version~\cite{Galinier99}. 
Checking that a 1-move is tabu or not and updating the tabu list are operations  performed in constant time. 
\emph{TabuCol} also uses an incremental evaluation~\cite{Fleurent96a}: 
the objective function of the neighbors is not computed from scratch, but only the difference between the two solutions is computed.
This is a very important feature for local search efficiency. 
Finding the best 1-move corresponds to find the maximum value of a $|V|\times k$ integer matrix. 
An efficient implementation of incremental data structures is well explained in~\cite{Galinier06}.

Another benefit of this version of \emph{TabuCol} is that it has only two parameters, $L$ and $\lambda$ to adjust in order to control the tabu tenure, $d$, by:
 $$d=L+\lambda F(s)$$ 
where $F(s)$ is the number of conflicting vertices in the curent solution $s$.
Moreover, \cite{Galinier99} has demonstrated on a very large number of instances that with the same setting 
($L$ a random integer inside $[0;\;9]$ and $\lambda=0.6$), \emph{TabuCol} obtained very good $k$-colorings.
Indeed, one of the main disadvantages of heuristics is that the number of parameters to set is high and difficult to adjust. 
This version of \emph{TabuCol} is very robust.
Thus we retained the setting of~\cite{Galinier99} in all our tests.


\subsection{Memetic Algorithms for graph coloring and \emph{HEA}}\label{sect:GPX}
Memetic Algorithms~\cite{Hao12} (MA) are hybrid metaheuristics using a local search algorithm inside a population-based algorithm. 
They can also be viewed as specific Evolutionary Algorithms (EAs) where all individuals of the population are local minimums (of a specific neighborhood).
In MA, the mutation of the EA is replaced by a local search algorithm.
It is very important to note that most of the running time of a MA is spent in the local search.
These hybridizations combine the benefits of 
population-based methods, which are better for diversification by means of a crossover operator, 
and local search methods, which are better for intensification.

In graph coloring, the Hybrid Evolutionary Algorithm (\emph{HEA}) of Galinier and Hao~\cite{Galinier99} is a MA;
the mutation of the EA is replaced by the tabu search \emph{TabuCol}.
\emph{HEA} is one of the best algorithms for solving the GVCP; 
From 1999 until 2012, it provided most of the best results for DIMACS benchmark graphs~\cite{dimacs96},
particularly for difficult graphs such as <\texttt{dsjc500.5}> and <\texttt{dsjc1000.5}> (see table~\ref{tab:synthesisColoring}).
These results were obtained with a population of 10 individuals.

%

\begin{figure}
\begin{center}
\includegraphics[width=0.9\linewidth]{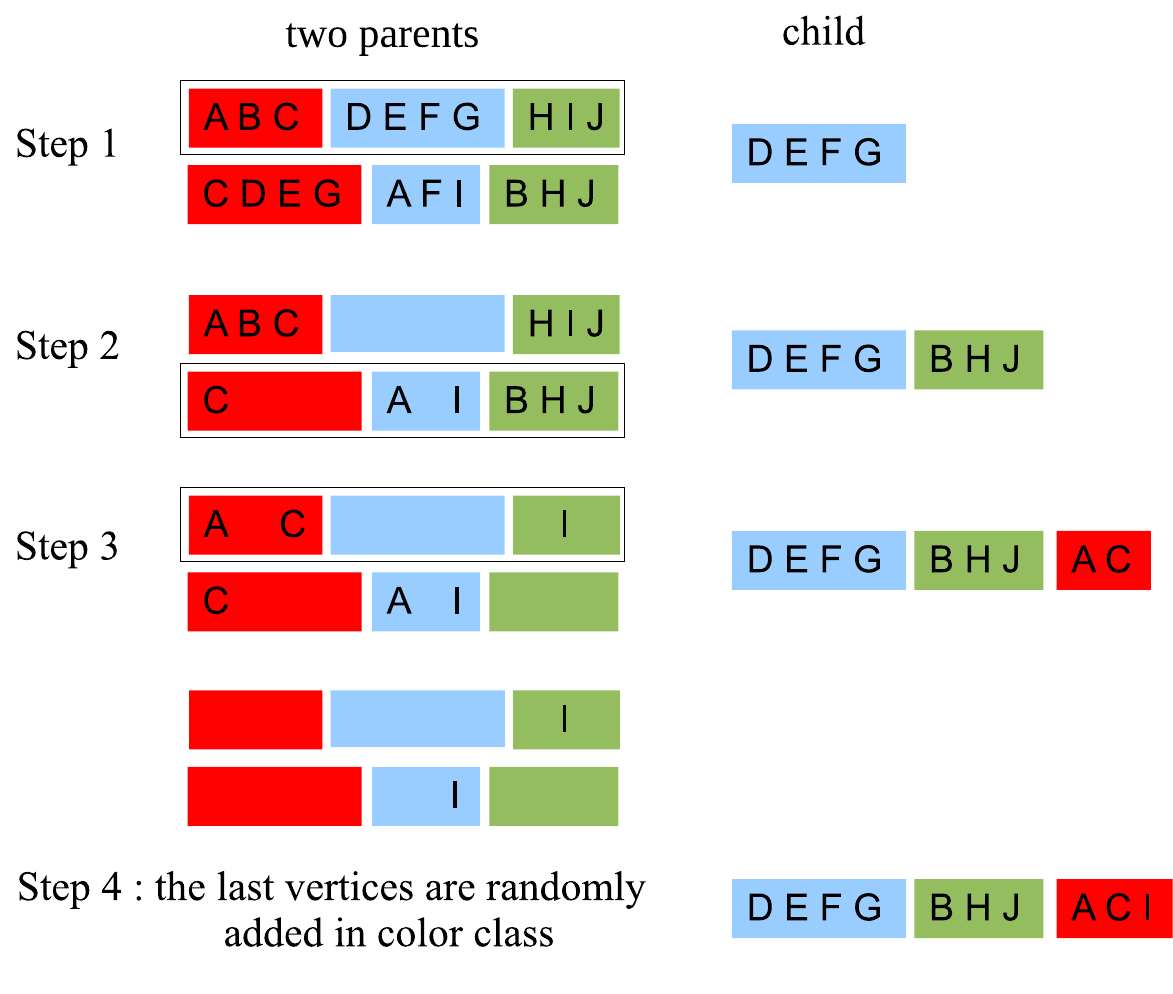}
\caption{\label{fig:GPX}An example of \emph{GPX} crossover for a graph of 10 vertices (A, B, C, D, E, F, G, H, I and J) and three colors (red, blue and green).
This example comes from~\cite{Galinier99}.}
\end{center}
\end{figure}

The crossover used in \emph{HEA} is called the Greedy Partition Crossover (\emph{GPX}). 
The two main principles of \emph{GPX} are: 1) a coloring is a partition of vertices into color classes and not an assignment of colors to vertices, and 
2) large color classes should be transmitted to the child.
Figure~\ref{fig:GPX} gives an example of \emph{GPX} for a problem with three colors (red, blue and green) and 10 vertices (A, B, C, D, E, F, G, H, I and J).
The first step is to transmit to the child the largest color class of the first parent. 
If there are several largest color classes, one of them is chosen at random. 
After having withdrawn those vertices in the second parent, one proceeds to step 2 where
one transmits to the child the largest color class of the second parent.
This process is repeated until all the colors are used.
There are most probably still some uncolored vertices in the child solution.
The final step (step $k+1$) is to randomly add those vertices to the color classes.
Notice that \emph{GPX} is asymmetrical: the order of the parents is important; starting the crossover with parent 1 or parent 2 can produce very different offsprings.
Notice also that \emph{GPX} is a random crossover: applying \emph{GPX} twice with the same parents does not produce the same offspring.
The final step is very important because it produces many conflicts. Indeed if the two parents have very different structures (in terms of color classes), then a large number of vertices remain uncolored at step $k+1$, and there are many conflicting edges in the offspring (cf. figure~\ref{fig:diversity_crossover}).
We investigate some modifications of \emph{GPX} in section~\ref{sect:analysis}.

\subsection{QA-col: Quantum Annealing for graph coloring}\label{ssec:qacol}
In 2012 Olawale Titiloye and Alan Crispin~\cite{Titiloye11a,Titiloye11b,Titiloye12} proposed a Quantum Annealing algorithm for graph coloring, denoted QA-col.
QA-col produces most of the best-known colorings for the DIMACS benchmark. In order to achieve this level of performance, QA-col is based on parallel computing. We briefly present some aspects of this new type of algorithm.

In a standard Simulating Annealing algorithm (SA), the probability of accepting a candidate solution is managed through a temperature criterion.
The value of the temperature decreases during the SA iterations.
As MA, a Quantum Annealing (QA) is a population-based algorithm, but it does not perform crossovers and the local search is an SA.
The only interaction between the individuals of the population occurs through a specific local attraction-repulsion process.
The SA used in QA-col algorithm is a $k$-fixed penalty strategy like \emph{TabuCol}: the individuals are non-proper $k$-colorings.
The objective function of each SA minimizes a linear combination of the number of conflicting edges and a given population diversity criterion as detailed later. 
The neighborhood used is defined by critic 1-moves like \emph{TabuCol}. 
More precisely, in QA-col, the $k$-colorings of the population are arbitrarily ordered in a ring topology: 
each $k$-coloring has two neighbors associated with it.
The second term of the objective function
(called Hamiltonian cf. equation (1) of \cite{Titiloye11a}) can be seen as a diversity criterion based on a  specific distance applicable to partitions. Given 
two $k$-colorings (i.e. partitions) $c_i$ and $c_j$, the distance, which we called \textit{pairwise partition distance}, between $c_i$ and $c_j$ is the following~: 
$$\displaystyle d_P(c_i,c_j)=\sum_{(u,v)\in V^2,\ u\neq v} [c_i(u)=c_i(v)]\oplus[c_j(u)=c_j(v)]$$ 
where $\oplus$ is the XOR operation 
and $[\ ]$ is Iverson bracket: $[c_i(u)=c_i(v)]=1$ if $c_i(u)=c_i(v)$ is true and equals $0$ otherwise.
Then, given one $k$-coloring $c_i$ of the population, the diversity criterion $D(c_i)$ is defined \supp{the distance is deduced of one $k$-coloring $c$ of the population to this population} as the sum of the \textit{pairwise partition distances} between $c_i$ and its two neighbors $c_{i+1}$ and $c_{i-1}$ in the ring topology~: 
$D(c_i)=d_P(c_i,c_{i-1}) + d_P(c_i,c_{i+1})$
which ranges from $0$ to $2n(n-1)$; 
The value of this diversity \supp{distance} is integrated into the objective function of each SA. 
As with the temperature, if the distance increases, there will be a higher probability that the solution will be accepted (\emph{attractive process}).
If the distance decreases, then there will be a lower probability that the solution will be accepted (\emph{repulsive process}).
Therefore in \emph{QA-col} the only interaction between the $k$-colorings of the population is realized through this distance process.

Although previous approaches are very efficient the reasons for this are difficult to assess.
They use many parameters and several intensification and diversification operators and thus the benefit of each item is not easily evaluated.
Our approach has been to identify which elements of \emph{HEA} are the most significant in order to define a more efficient algorithm.


\section{\emph{HEAD}: Hybrid Evolutionary Algorithm in Duet}\label{sect:HEAD}

The basic components of \emph{HEA} are the \emph{TabuCol} algorithm, which is a very powerful local search for intensification, 
and the \emph{GPX} crossover, which adds some diversity.
The intensification/diversification balance is difficult to achieve.
In order to simplify the numerous parameters involved in EAs, we have chosen to consider a population with only two individuals.
We present two versions of our algorithm denoted \emph{HEAD'} and \emph{HEAD} for \emph{HEA} in Duet.

\subsection{First hybrid algorithm: HEAD'}
Algorithm~\ref{alg:algo1} describes the pseudo code of the first version of the proposed algorithm, 
denoted \emph{HEAD'}. 
\begin{algorithm}
\footnotesize
\caption{- \emph{HEAD'} - first version of \emph{HEAD}: \emph{HEA} in Duet for $k$-coloring problem}
\label{alg:algo1}
\DontPrintSemicolon
  \KwIn{$k$, the number of colors; $Iter_{TC}$, the number of \emph{TabuCol} iterations.}
  \KwOut{the best $k$-coloring found: $best$}
$p_1 , p_2, best \leftarrow $ init()\tcc*{initialize with random $k$-colorings} 
$generation \leftarrow 0$

  \Do{$nbConflicts(best)>0$ and $p_1\neq p_2$}{
    $c_1 \leftarrow $ \emph{GPX}($p_1$, $p_2$)\;
    $c_2 \leftarrow $ \emph{GPX}($p_2$, $p_1$)\;
    $p_1 \leftarrow $ \emph{TabuCol}($c_1$,$Iter_{TC}$)\;
    $p_2 \leftarrow $ \emph{TabuCol}($c_2$,$Iter_{TC}$)\;
    $best\leftarrow$ saveBest($p_1,p_2,best$)\;
  $generation++$
  }
\end{algorithm}
This algorithm can be seen as 
two parallel \emph{TabuCol} algorithms which periodically interact by crossover.


%
%
%

After randomly initializing the two solutions (with init() function), the algorithm repeats an instructions loop until a stop criterion occurs.
First, we introduce some diversity with the crossover operator, then the two offspring $c_1$ and $c_2$ are improved by means of the \emph{TabuCol} algorithm.
Next, we register the best solution and we systematically replace the parents by the two children.
An iteration of this algorithm is called a \emph{generation}.
\supp{The majority of the running time of a generation is spent in the \emph{TabuCol} algorithm.}
The main parameter of \emph{TabuCol} is $Iter_{TC}$, the number of iterations performed by the algorithm, 
the other \emph{TabuCol} parameters $L$ and $\lambda$ are used to define the tabu tenure and are considered fixed in our algorithm.
Algorithm~\ref{alg:algo1} stops either because a legal $k$-coloring is found ($nbConflicts(best)=0$) 
or because the two $k$-colorings are equal in terms of the \emph{set-theoretic partition distance
(\emph{cf.} Section}~\ref{sect:analysis}).

A major risk is a premature convergence of \emph{HEAD'}. Algorithm~\ref{alg:algo1} stops sometimes too quickly: the two individuals are equal before finding a legal coloring. It is then necessary to reintroduce diversity into the population. In conventional EAs, the search space exploration is largely brought by the size of the population: the greater the size, the greater the search diversity. In the next section we propose an alternative to the population size in order to reinforce diversification.


%
%
%
%
%
\subsection{Improved hybrid algorithm: HEAD}

\begin{algorithm}
\footnotesize
  \DontPrintSemicolon
  \KwIn{$k$, the number of colors; $Iter_{TC}$, the number of \emph{TabuCol} iterations; $Iter_{cycle}=10$, the number of generations into one cycle.}
  \KwOut{the best $k$-coloring  found: $best$}
$p_1 , p_2, elite_1, elite_2, best \leftarrow $ init()\tcc*{initialize with random $k$-colorings} 
 $generation,\ cycle \leftarrow 0$

  \Do{$nbConflicts(best)>0$ and $p_1\neq p_2$}{
    $c_1 \leftarrow $ \emph{GPX}($p_1$, $p_2$)\;
    $c_2 \leftarrow $ \emph{GPX}($p_2$, $p_1$)\;
    $p_1 \leftarrow $ \emph{TabuCol}($c_1$,$Iter_{TC}$)\;
    $p_2 \leftarrow $ \emph{TabuCol}($c_2$,$Iter_{TC}$)\;
    $elite_1\leftarrow$ saveBest($p_1$, $p_2$, $elite_1$)\tcc*{best $k$-coloring of the current cycle}
    $best\leftarrow$ saveBest($elite_1, best$)\;
     \If{$generation\%Iter_{cycle}=0$}{
       $p_1 \leftarrow elite_2$\tcc*{best $k$-coloring of the previous cycle}
       $elite_2 \leftarrow elite_1$\;
       $elite_1 \leftarrow$ init()\;
       $cycle++$\;
   }
  $generation++$\;
  }
 \caption{\emph{HEAD} - second version of \emph{HEAD} with two extra elite solutions}
 \label{alg:algo2}
\end{algorithm}

\begin{figure}
\begin{center}
 \includegraphics[width=.6\linewidth]{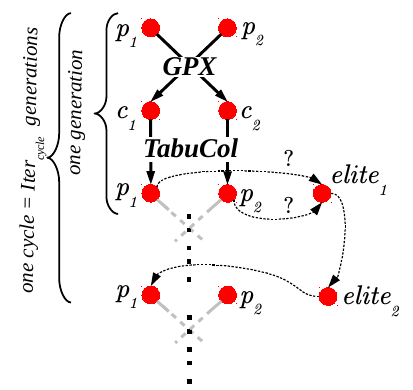}  
 \caption{\label{fig:H2}Diagram of \emph{HEAD}}
\end{center}
 \end{figure}

Algorithm~\ref{alg:algo2} summarizes the second version of our algorithm, simply denoted \emph{HEAD}.
We add two other candidate solutions (similar to elite solutions), $elite_1$ and $elite_2$, 
in order to reintroduce some diversity to the duet. 
Indeed, after a given number of generations, 
the two individuals of the population become increasingly similar within the search-space.
To maintain the population diversity, the idea is to replace one of the two candidates solutions by a solution previously encountered by the algorithm. We define one cycle as a number of $Iter_{cycle}$ generations. Solution $elite_1$ is the best solution found during the current cycle and solution $elite_2$ the best solution found during the previous cycle. At the end of each cycle, the $elite_2$ solution 
replaces one of the population individuals.
Figure~\ref{fig:H2} presents the graphic view of algorithm~\ref{alg:algo2}. 

 {
This elitist mechanism provides relevant behaviors to the algorithm as it can be observed in the computational results of section~\ref{ssec:results}. Indeed, elite solutions have the best fitness value of each cycle. It is clearly interesting in terms of intensification. Moreover, when the elite solution is reintroduced, it is generally different enough from the other individuals to be relevant in terms of diversification. 
In the next section, we show how the use of this elitist mechanism can enhance the results.







\section{Experimental Results}\label{sect:results}
In this section we present the results obtained with the two versions of the proposed memetic algorithm. 
To validate the proposed approach, the results of \emph{HEAD} are compared with the results obtained by the best methods currently known. 

\subsection{Instances and Benchmarks}
Test instances are selected among the most studied graphs since the 1990s, which are known to be very difficult (the second DIMACS challenge of 1992-1993~\cite{dimacs96}\footnote{\texttt{ftp://dimacs.rutgers.edu/pub/challenge/graph/\\benchmarks/color/}}).

We focus during the study on some types of graphs from the DIMACS benchmark: <\texttt{dsjc}>, <\texttt{dsjr}>, <\texttt{flat}>, <\texttt{r}>, <\texttt{le}> and <\texttt{C}> 
which are randomly or quasi-randomly generated graphs.
<\texttt{dsjc}$n.d$> graphs and <\texttt{C}$n.d$> graphs are random graphs with $n$ vertices, 
with each vertex connected to an average of $n\times 0.d$ vertices; $0.d$ is the graph density. 
The chromatic number of these graphs is unknown. 
Likewise for <\texttt{r}$n.d$[\texttt{c}]> and <\texttt{dsjr}$n.d$> graphs which are geometric random graphs 
with $n$ vertices and a density equal to $0.d$. [\texttt{c}] denotes the complement of such a graph.
<\texttt{flat}> and <\texttt{le}> graphs have another structure: 
they are built for a known chromatic number. 
The <\texttt{flat}$n\_\chi$> graph or <\texttt{le}$n\_\chi$[\texttt{abcd}]> graph has $n$ vertices and $\chi$ is the chromatic number.

\subsection{Computational Results} \label{ssec:results} 
\emph{HEAD} and \emph{HEAD'} were coded in C++. 
The results were obtained with an Intel Core i5 3.30GHz processor - 4 cores and 16GB of RAM. 
Note that the RAM size has no impact on the calculations: even for large graphs such as <\texttt{dsjc1000.9}> (with 1000 vertices and a high density of 0.9), 
the use of memory does not exceed 125 MB. The main characteristic is the processor speed.

As shown in Section~\ref{sect:HEAD}, the proposed algorithms have two successive calls to local search 
(lines 6 and 7 of the algorithms~\ref{alg:algo1} and~\ref{alg:algo2}), 
one for each child of the current generation. Almost all of the time is spent on performing the local search. 
Both local searches can be parallelized when using a multi-core processor architecture. 
This is what we have done using the OpenMP API (Open Multi-Processing), 
which has the advantage of being a cross-platform (Linux, Windows, MacOS, etc.) and simple to use. 
Thus, when an execution of 15 minutes is given, the required CPU time is actually close to 30 minutes if using only one processing core.

Table~\ref{tab:synthesisColoring} presents results of the principal methods known to date for 19 \textit{difficult} graphs. 
For each graph, the lowest number of colors found by each algorithm is indicated (upper bound of $\chi$). 
For \emph{TabuCol} \cite{Hertz87} the reported results are from~\cite{Hertz08} (2008) which are better than those of 1987.
The most recent algorithms, \emph{QA-col} (Quantum Annealing for graph coloring~\cite{Titiloye12}) and \emph{IE$^2$COL} (Improving the Extraction and Expansion
method for large graph COLoring~\cite{Hao12b}), 
provide the best results but \emph{QA-col} is based on a cluster of PC using 10 processing cores simultaneously and \emph{IE$^2$COL} is profiled for large graphs ($>900$ vertices).
Note that \emph{HEA}~\cite{Galinier99}, \emph{AmaCol}~\cite{Galinier08}, \emph{MACOL}~\cite{Lu10}, \emph{EXTRACOL}~\cite{Wu12} and \emph{IE$^2$COL} are also population-based algorithms 
using \emph{TabuCol} and \emph{GPX} crossover or an improvement of \emph{GPX} (\emph{GPX} with $n\geqslant2$ parents for \emph{MACOL} and \emph{EXTRACOL} and 
the \emph{GPX} process is replaced in \emph{AmaCol} by a selection of $k$ color classes among a very large pool of color classes). 
Only \emph{QA-col} has another approach based on several parallel simulated annealing algorithms interacting together with an attractive/repulsive process (cf. section~\ref{ssec:qacol}). 

\begin{table*}[htb]
    \begin{center}
	\rowcolors{5}{white}{blue!4}
	\scriptsize
	\begin{tabular}{|c||c||c||cccccc|}
		\hline 
		&  & LS & \multicolumn{6}{c|}{Hybrid algorithm}\\
		\cline{3-9} 
		&  & 1987/2008 & 1999 & 2008 & 2010 & 2011 & 2012 & 2012\\
		Graphs & \emph{HEAD} & \emph{TabuCol}& \emph{HEA}& AmaCol& MACOL& EXTRACOL& IE$^2$COL& QA-col\\  
		&&\cite{Hertz87,Hertz08}&\cite{Galinier99}&\cite{Galinier08}&\cite{Lu10}&\cite{Wu12}&\cite{Hao12b}& \cite{Titiloye12}\\ \hline \hline
		\texttt{dsjc250.5} & \textbf{28} & \textbf{28} & \textbf{28} & \textbf{28} & \textbf{28} & - &- & \textbf{28} \\
		\texttt{dsjc500.1} & \textbf{12} & 13 & - & \textbf{12}  & \textbf{12}  & - & - & -\\
		\texttt{dsjc500.5} & \textbf{47} & 49 & 48 & 48 & 48 & - & - & \textbf{47}\\
		\texttt{dsjc500.9} & \textbf{126} & 127 & -  & \textbf{126}  & \textbf{126}  & - & -  & \textbf{126} \\
		\texttt{dsjc1000.1} & \textbf{20} & - & \textbf{20} & \textbf{20} & \textbf{20} & \textbf{20} & \textbf{20} & \textbf{20}\\
		\texttt{dsjc1000.5} & \textbf{82} & 89 & 83 & 84 & 83 & 83 & 83 & \textbf{82}\\
		\texttt{dsjc1000.9} & \textbf{222} & 227 & 224 & 224 & 223 & \textbf{222} & \textbf{222} & \textbf{222}\\ 
		\texttt{r250.5} & \textbf{65} & - & -  & - & \textbf{65} & -  & -  & \textbf{65}\\
		\texttt{r1000.1c} & \textbf{98} & - & -  & - & \textbf{98} & 101  & \textbf{98} & \textbf{98} \\
		\texttt{r1000.5} & 245 & - & -  & - & 245 & 249  & 245 & \textbf{234} \\
		\texttt{dsjr500.1c} & \textbf{85} & \textbf{85} & -  & 86 & \textbf{85} & -  & -  & \textbf{85} \\ 
		\texttt{le450\_25c} & \textbf{25} & 26 & 26  & 26 & \textbf{25} & -  & - & \textbf{25} \\
		\texttt{le450\_25d} & \textbf{25} & 26 & -  & 26 & \textbf{25} & -  & - & \textbf{25} \\
		\texttt{flat300\_28\_0} & 31 & 31 & 31  & 31 & \textbf{29} & -   & - & 31 \\
		\texttt{flat1000\_50\_0} & \textbf{50} & \textbf{50} & -  & \textbf{50}  & \textbf{50} & \textbf{50} & \textbf{50}  & - \\
		\texttt{flat1000\_60\_0} & \textbf{60} & \textbf{60}  & -  & \textbf{60} & \textbf{60}  & \textbf{60}  & \textbf{60}  & - \\
		\texttt{flat1000\_76\_0} & \textbf{81} & 88 & 83 & 84 & 82 & 82 & \textbf{81}& \textbf{81}\\ 
		\texttt{C2000.5} & 146 & - & -  & - & 148 & 146 & \textbf{145}  & \textbf{145}   \\
		\texttt{C4000.5} & 266 & - & -  & - & 272 & 260 & \textbf{259} & \textbf{259}   \\
		\hline 
	\end{tabular}
\hspace{20mm} 
\end{center}
\caption{\label{tab:synthesisColoring}Best coloring found}
\end{table*}

Table~\ref{tab:resSimpleH2col} presents the results obtained with \emph{HEAD'}, the first version of \emph{HEAD} (without elite). 
This simplest version finds the best known results for most of the studied graphs (13/19);
Only QA-col (and IE$^2$COL for <\texttt{C}> graphs) occasionally finds a solution with less color.
The column $\mathbf{Iter_{TC}}$ indicates the number of iterations of the \emph{TabuCol} algorithm 
(this is the stop criterion of \emph{TabuCol}). This parameter has been determined for each graph after an empirical analysis for finding the most suitable value.
The column \textbf{GPX} refers to the \emph{GPX} used inside \emph{HEAD'}. Indeed, in section~\ref{sect:analysis}, we define two modifications of the standard \emph{GPX} (Std): 
the unbalanced \emph{GPX} (U($\intervalleff{0}{1}$)) and the random \emph{GPX} (R($\iintervalleff{0}{k}$)).
One can notice that the choice of the unbalanced or the random crossover is based on the study of the algorithm in the standard mode (standard \emph{GPX}). If the algorithm needs too many generations for converging we introduce the unbalanced \emph{GPX}. At the opposite, if the algorithm converges quickly without finding any legal k-coloring we introduce the random crossover.  
Section~\ref{sect:analysis} details the modifications of the \emph{GPX} crossover 
(section~\ref{subsect:GPX_random} for the random \emph{GPX} and section~\ref{subsect:GPX_balence} for the unbalanced \emph{GPX}).

The column \textbf{Success} evaluates the robustness of this method, providing the success rate: success\_runs/total\_runs. 
A success run is one which finds a legal $k$-coloring. 
The average number of generations or crossovers performed during one success run is given by the~\textbf{Cross} value.
The total average number of iterations of \emph{TabuCol} preformed during \emph{HEAD'} is $$\mathbf{Iter=Iter_{TC}\times Cross}\times 2.$$
The column \textbf{Time} indicates the average CPU time in minutes of success runs. 

\supp{\emph{HEAD'} does not find the solutions each time for these graphs, but when it does, it is generally very fast. }
\emph{HEAD'} success rate is rarely 100\%, but in case of success, the running time is generally very short.
The main drawback of \emph{HEAD'} is that it sometimes converges too quickly. 
In such instances it cannot find a legal solution before the two individuals in a generation become identical. 
The first option to correct this rapid convergence, is to increase the number of iterations $Iter_{TC}$ of each \emph{TabuCol}. The second option is to use the random \emph{GPX} instead of the standard one (section~\ref{subsect:GPX_random}). 
However, these options are not considered sufficient.
The second version, \emph{HEAD}, adds more diversity while performing an intensifying role.

\begin{table*}[htb]
    \begin{center}
	\centering
	\rowcolors{2}{white}{blue!4}
	\scriptsize
	\begin{tabular}{|c|ccccccc|}
		\hline 
		\textbf{Instances}&\textbf{k}&$\mathbf{Iter_{TC}}$&\textbf{GPX}&\textbf{Success}&\textbf{Iter}&\textbf{Cross}&\textbf{Time} \\ \hline
		\texttt{dsjc250.5} 		& \textbf{28}	& 6000		& Std	& 17/20	& $1\times10^6$		& 79	& 0.01 min	\\  \hline
		\texttt{dsjc500.1} 		& \textbf{12}	& 8000		& Std	& 15/20	& $2.5\times10^6$	& 158	& 0.03 min	\\ \hline 
		\texttt{dsjc500.5} 		& 48	 		& 8000		& Std	& 9/20	& $5.3\times10^6$	& 334	& 0.2 min	\\ \hline 
		\texttt{dsjc500.9} 		& \textbf{126}	& 25000		& Std	& 10/20	& $2.5\times10^7$	& 517	& 1 min		\\ \hline 
		\texttt{dsjc1000.1} 		& \textbf{20}	& 7000		& Std	& 7/20	& $8.2\times10^6$	& 588	& 0.2 min	\\ \hline  
		\texttt{dsjc1000.5} 		& 83 			& 40000		& Std	& 16/20	& $1.37\times10^8$	& 1723	& 10 min	\\ \hline  
		\texttt{dsjc1000.9} 		& \textbf{222} 	& 60000		& Std	& 1/20	& $4.45\times10^8$	& 3711	& 33 min	\\
						& 223 			& 30000		& Std	& 4/20	& $6.6\times10^7$ 	& 1114	& 5 min		\\ \hline  
		
		\texttt{r250.5}			& \textbf{65} 	& 12000		& Std	& 1/20	& $8.11\times10^8$ 	& 33828	& 12 min	\\ 
						& \textbf{65} 	& 2000		& R(20)	& 6/20	& $5.31\times10^8$ 	& 132773& 10 min	\\ \hline
						
		\texttt{r1000.1c}		& \textbf{98} 	& 65000		& Std	& 1/20	& $2.32\times10^6$ 	& 18	& 0.1 min	\\
						& \textbf{98} 	& 25000		& R(98) & 20/20	& $6.5\times10^6$ 	& 130	& 0.4 min	\\ \hline
						
		\texttt{r1000.5}			& 245		 	& 360000	& Std 	& 20/20	& $2.6\times10^9$ 	& 3636	& 135 min	\\
						& 245		 	& 240000	& U(0.98)& 17/20& $6.48\times10^8$ 	& 1352	& 39 min	\\ \hline
		
		\texttt{dsjr500.1c}		& \textbf{85}	& 4200000	& Std	& 1/20	& $5.8\times10^6$	& 1		& 0.2 min	\\ 
						& \textbf{85} 	& 1000		& R(85)& 13/20	& $5\times10^5$	& 279	& 0.02 min	\\ \hline

		\texttt{le450\_25c}		& \textbf{25} 	& 21000000	& Std	& 20/20	& $3.5\times10^9$ 	& 57	& 38 min	\\ 
						& \textbf{25} 	& 300000	& U(0.98)& 10/20& $2.86\times10^8$ 	& 477	& 2.4 min	\\ \hline
		\texttt{le450\_25d}		& \textbf{25} 	& 21000000	& Std 	& 20/20	& $5.7\times10^9$ 	& 135	& 64 min	\\ 
						& \textbf{25} 	& 340000	& U(0.98)& 10/20& $2.15\times10^8$ 	& 317	& 2 min		\\ \hline
		
		\texttt{flat300\_28\_0} 	& 31			& 4000		& Std	& 20/20	& $1\times10^6$		& 117	& 0.02 min	\\ \hline 
		\texttt{flat1000\_50\_0} & \textbf{50}	& 130000	& Std	& 20/20	& $1.1\times10^6$	& 4		& 0.3 min	\\ \hline 
		\texttt{flat1000\_60\_0} & \textbf{60}	& 130000	& Std	& 20/20	& $2.3\times10^6$	& 9		& 0.5 min	\\ \hline 
		\texttt{flat1000\_76\_0} & \textbf{81} 	& 40000		& Std	& 1/20	& $1.49\times10^9$	& 18577	& 137 min	\\
						& 82 			& 40000		& Std	& 18/20	& $1.57\times 10^8$	& 1969	& 11 min	\\ \hline 
        \texttt{C2000.5}			& 148		 	& 140000& Std	& 10/10	& $1.7\times10^9$	& 6308		& 794 min	\\	\hline		
		\texttt{C4000.5} 		& 275			& 140000& Std	& 8/10	& $1.1\times10^9$	& 4091		& 3496 min	\\ 	\hline
	\end{tabular}
   \end{center}
	\caption{\label{tab:resSimpleH2col}Results of \emph{HEAD'}, the first version of \emph{HEAD} algorithm (without elites)}
\end{table*}

Table~\ref{tab:resH2col} shows the results obtained with \emph{HEAD}. 
For all the studied graphs except four (<\texttt{flat300\_28\_0}>, <\texttt{r1000.5}>, <\texttt{C2000.5}> and <\texttt{C4000.5}>), \emph{HEAD} finds the best known results.
Only the Quantum Annealing algorithm, using ten CPU cores simultaneously, and \emph{IE$^2$COL} for large graphs, achieve this level of performance. 
In particular, <\texttt{dsjc500.5}> is solved with only 47 colors and <\texttt{flat1000\_76\_0}> with 81 colors. 

\begin{table*}[htb]
    \begin{center}
	\centering
	\rowcolors{2}{white}{blue!4}
	\scriptsize
	\begin{tabular}{|c|ccccccc|}
		\hline 
		\textbf{Instances}& \textbf{k}	&$\mathbf{Iter_{TC}}$& \textbf{GPX} &\textbf{Success} & \textbf{Iter} 	& \textbf{Cross}	& \textbf{Time}	\\ \hline
		\texttt{dsjc250.5}		& \textbf{28}	& 6000	& Std	& 20/20	& $9\times10^5$	& 77		& 0.01 min	\\ \hline
		\texttt{dsjc500.1}		& \textbf{12}	& 4000	& Std	& 20/20 & $3.8\times10^6$	& 483		& 0.1 min	\\ \hline
		\texttt{dsjc500.5}		& \textbf{47}	& 8000	& Std	& 2/10000& $2.4\times10^7$	& 1517		& 0.8 min	\\
		& 48 			& 8000	& Std	& 20/20	& $7.6\times10^6$	& 479		& 0.2 min	\\ \hline
		\texttt{dsjc500.9} 		& \textbf{126}	& 15000	& Std	& 13/20	& $2.9\times10^7$	& 970		& 1.2 min	\\ \hline
		\texttt{dsjc1000.1} 		& \textbf{20} 	& 3000	& Std	& 20/20	& $3.4\times10^6$	& 567		& 0.2 min	\\ \hline
		\texttt{dsjc1000.5}		& \textbf{82}	& 60000	& Std	& 3/20	& $1\times10^9$		& 8366 		& 48 min	\\
		& 83 			& 40000	& Std	& 20/20	& $9.6\times10^7$	& 1200		& 6 min		\\ \hline
		\texttt{dsjc1000.9} 		& \textbf{222} 	& 50000	& Std	& 2/20	& $1.2\times10^9$	& 11662		& 86 min	\\    
		& 223   		& 30000	& Std	& 19/20	& $1.26\times10^8$	& 2107		& 10 min	\\ \hline  
		
		\texttt{r250.5} 			& \textbf{65} 	& 10000	& Std	& 1/20	& $6.98\times10^8$	& 34898		& 13 min	\\
			 			& \textbf{65} 	& 4000	& R(20)	& 20/20	& $3.91\times10^8$	& 48918		& 6.3 min	\\ \hline
		\texttt{r1000.1c}	 	& \textbf{98}	& 45000	& Std	& 3/20	& $3.7\times10^6$	& 42		& 0.2 min	\\
						& \textbf{98}  	& 25000	& R(98)	& 20/20	& $3.9\times10^6$	& 78		& 0.24 min	\\ \hline
		\texttt{r1000.5} 		& 245   		& 360000& Std	& 20/20	& $4.6\times10^9$	& 6491		& 244 min	\\
						& 245   		& 240000& U(0.98)& 20/20& $5.3\times10^8$	& 1104		& 25 min	\\ \hline 
		
		\texttt{dsjr500.1c} 		& \textbf{85}	& 4200000 & Std	& 1/20	& $5.8\times10^6$	& 1		& 0.2 min	\\ 
						& \textbf{85} 	& 400	& R(85)	& 20/20	& $4\times10^5$	& 408		& 0.02 min	\\ \hline  
		
		\texttt{le450\_25c} 		& \textbf{25}  	& 22000000& Std	& 20/20	& $2.7\times10^9$	& 62		& 30 min	\\
						& \textbf{25}	& 220000& U(0.98)& 20/20& $3.89\times10^8$	& 885		& 5 min		\\	 \hline  			
		\texttt{le450\_25d}		& \textbf{25}  	& 21000000& Std	& 20/20	& $7\times10^9$	& 161		& 90 min	\\  				
						& \textbf{25}  	& 220000& U(0.98)& 20/20& $2.35\times10^8$	& 534		& 2 min		\\ \hline
		
		\texttt{flat300\_28\_0} 	& 31 			& 4000	& Std	& 20/20	& $9\times10^5$	& 120		& 0.02 min	\\ \hline
		\texttt{flat1000\_50\_0} & \textbf{50}	& 130000& Std	& 20/20	& $1.2\times10^6$	& 5			& 0.3 min	\\ \hline 
		\texttt{flat1000\_60\_0} & \textbf{60}	& 130000& Std	& 20/20	& $2.3\times10^6$	& 9			& 0.5 min	\\ \hline 			
		\texttt{flat1000\_76\_0} & \textbf{81} 	& 60000	& Std	& 3/20	& $1\times10^9$	& 8795		& 60 min	\\
						& 82	 		& 40000	& Std	& 20/20	& $8.4\times10^7$	& 1052		& 5 min		\\ \hline 
		\texttt{C2000.5}			& 146		 	& 140000& Std	& 8/10	& $1.78\times10^9$	& 6358		& 281 min	\\		
						& 147		 	& 140000& Std	& 10/10	& $7.26\times10^8$	& 2595		& 124 min	\\	\hline	
		\texttt{C4000.5} 	& 266			& 140000& Std	& 4/10	& $2.5\times10^9$	& 9034		& 1923 min	\\
        						& 267			& 140000& Std	& 8/10	& $1.6\times10^9$	& 5723		& 1433 min	\\ 	\hline	
	\end{tabular}
    \end{center}
	\caption{\label{tab:resH2col}Results of the second version of \emph{HEAD} algorithm (with elites) including the indication of CPU time}
\end{table*}

The computation time of \emph{HEAD} is generally close to that of \emph{HEAD'} but the former algorithm is more robust with a success rate of almost 100\%. 
In particular, the two graphs <\texttt{dsjc500.5}> and <\texttt{dsjc1000.1}> with 48 and 20 colors respectively are resolved each time, and in less than one CPU minute on average (CPU 3.3GHz). 
Using a multicore CPU, these instances are solved in less than 30 seconds on average, often in less than 10 seconds. 
As a comparison, the shortest time reported in the literature for <\texttt{dsjc1000.1}> is 32 minutes for \emph{QA-col}~\cite{Titiloye11b} (2011) with a 3GHz processor, 65 minutes for \emph{IE$^2$COL} (2012) with a 2.8GHz processor, 93 minutes for \emph{EXTRACOL}~\cite{Wu12} (2011) with a 2.8GHz processor and 108 minutes for \emph{MACOL}~\cite{Lu10} (2010) with a 3.4GHz processor.


\section{Analysis of diversification}\label{sect:analysis}

\emph{HEAD} shares common features with \emph{HEA}, but it obtains significantly better results with respect to solution quality and computing time. 
It is beneficial to analyze why the new mechanisms introduced with \emph{HEAD} gives rise to such a large change. 

A first answer can be formulated with regard to computing time. 
It can be observed that $99\%$ of the running time of \emph{HEA}, \emph{Amacol}, \emph{MACOL} and \emph{HEAD} is spent during calculating \emph{TabuCol} algorithms.
Considering a population of $10$ $k$-colorings in the case of \emph{HEA} and \emph{Amacol} ($20$ in case of \emph{MACOL}) requires more time than only two such individuals for \emph{HEAD}.

In our study \emph{HEAD} is not considered as a standard MA, but rather as two separated \emph{TabuCol} algorithms.
After a given number of iterations, instead of stopping the two \emph{TabuCol}, 
we reintroduce diversity with the crossover operator \emph{GPX}. 
The difficulty is to reintroduce the correct dose of diversity. 
Indeed the danger of the crossover is that of completely destroying the solution structure.
\emph{GPX} is a powerful crossover operator compared to others~\cite{Fleurent96a} 
because it transmits the biggest color classes of the two parents, thus keeping a large part of the parents' structures. Very interesting and relevant studies about how to manage diversity for graph coloring heuristics can be found in~\cite{Porumbel10b, Lewis2015}.

We present in this section an analysis of \emph{GPX} crossover 
indicating that it is more accurate to have parents that are not too far away in the search-space - according to the distance presented below (section~\ref{subsect:GPX_distance}). 

Several tests are also performed in this section in order to analyze the role of diversification in the \emph{HEAD} algorithm. 
The two main mechanisms leading to diversification in \emph{HEAD} are the \emph{GPX} crossover and the population update process.
In a first set of tests (section~\ref{subsect:GPX_modification}), we slightly modify the dose of diversification in the \emph{GPX} crossover and analyze the results. 
In a second set of tests (section~\ref{subsect:parent_replacement}), we focus on the population update process:
in \emph{HEAD}, the two produced children systematically replace both parents, even if they have worse fitness values than their parents. 
If the replacement is not systematic, the diversification decreases.

\subsection{Distance between parents and GPX crossover}\label{subsect:GPX_distance}

\emph{GPX} crossover is a diversification operator: it generates solutions in numerous uncharted regions of the search-space.
However, there is a risk of providing too much diversity, and thus breaking the structure of the current solution.
This is the principal failing of basic crossovers used before \emph{GPX}~\cite{Fleurent96a}.

An interesting feature of \emph{GPX} is its ability to explore new areas 
of the search-space without breaking the structures of current $k$-colorings.
There are many parameters that affect the dose of diversity of \emph{GPX}.
One of the easily identifiable parameters is the distance between the two parents. 

The \textit{set-theoretic partition distance}~\cite{Galinier99,Gusfield2002}
between two $k$-colorings $c_1$ and $c_2$ 
is defined as the least number of 1-move steps (i.e. a color change of one vertex) for transforming $c_1$ to $c_2$.
This distance has to be independent of the permutation of the color classes, then before counting the number of 1-moves,
we have to match each color class of $c_1$ with the nearest color class of $c_2$. 
This problem is a maximum weighted bipartite matching if we consider each color class of $c_1$ and $c_2$ as the vertices of a bipartite graph;
an edge links a color class of $c_1$ with a color class of $c_2$ with an associated value corresponding to the number of vertices shared by those classes.
The \textit{set-theoretic partition distance} is then calculated as follows: $d_H(c_1,c_2)=n-q$ where $n$ is the number of vertices of the initial graph and $q$ the result of the matching; 
i.e. the maximal total number of sharing vertices in the same class for $c_1$ and $c_2$. 
Figure~\ref{fig:dist} gives an example of the computation of this distance between two $3$-colorings.
The possible values range from $0$ to less than $n$. Indeed it is not possible to have totally different $k$-coloring.

 \begin{figure}[h!]
\begin{center}
 \includegraphics[width=.7\linewidth]{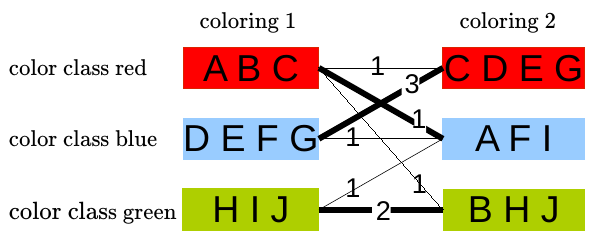}  
 \caption{\label{fig:dist}A graph with 10 vertices (A, B, C, D, E, F, G, H, I and J), three colors (red, blue and green) and
two $3$-colorings: coloring 1 and coloring 2. 
We defined the weighted bipartite graph corresponding to  
the number of vertices shared by color classes of coloring 1 and coloring 2. 
The bold lines correspond to the maximum weighted bipartite matching. The maximal total number of sharing vertices in the same class is equal to $q=3+2+1=6$.
Then the \emph{set-theoretic partition distance} between those two $3$-colorings is equal to: $d_H($coloring 1, coloring~2$)=n-q=10-6=4$. 
This distance is independent of the permutation of the color classes.}
\end{center}
 \end{figure}

If we highlight two $k$-colorings that have very low objective functions but that are very different (in terms of the $d_H$ distance),
then they would have a high probability of producing children with very high objective functions following crossover. 
The danger of the crossover is of completely destroying the $k$-coloring structure.
On the other hand, two very close $k$-colorings (in terms of the $d_H$ distance) produce a child with an almost identical objective function.
Chart~\ref{fig:diversity_crossover} shows the correlation between 
the distance separating two $k$-colorings having the same number of conflicting edges (objective functions equal to 40)
and the number of conflicting edges of the child produced after \emph{GPX} crossover.
This chart is obtained considering $k=48$ colors into the <\texttt{dsjc500.5}> graph. 
More precisely, this chart results of the following steps:
1) First, 100 non legal $48$-colorings, called parents, are randomly generated with a fitness (that is a number of conflicting edges) equal to 40. \emph{Tabucol} algorithm is used to generate these 100 parents (\emph{Tabucol} is stopped when exactly 40 conflicted edges are found).
2) A \emph{GPX} crossover is performed on all possible pairs of parents, generating for each pair two new non legal $48$-colorings, called children. Indeed, \emph{GPX} is asymmetrical, then the order of the parents is important. By this way, $9900$ ($=100\times99$) children are generated.
3) We perform twice the steps 1) and 2), therefore the total number of generated children is equal to  $19800$. Each point of the chart corresponds to one child.
The $y$-axis indicates the fitness of the child.
The $x$-axis indicates the distance $d_H$ in the search-space between the two parents of the child.
There is a quasi-linear correlation between these two parameters (Pearson correlation coefficient equals to 0.973). 
 \begin{figure}[h]
\begin{center}
 \includegraphics[width=.9\linewidth]{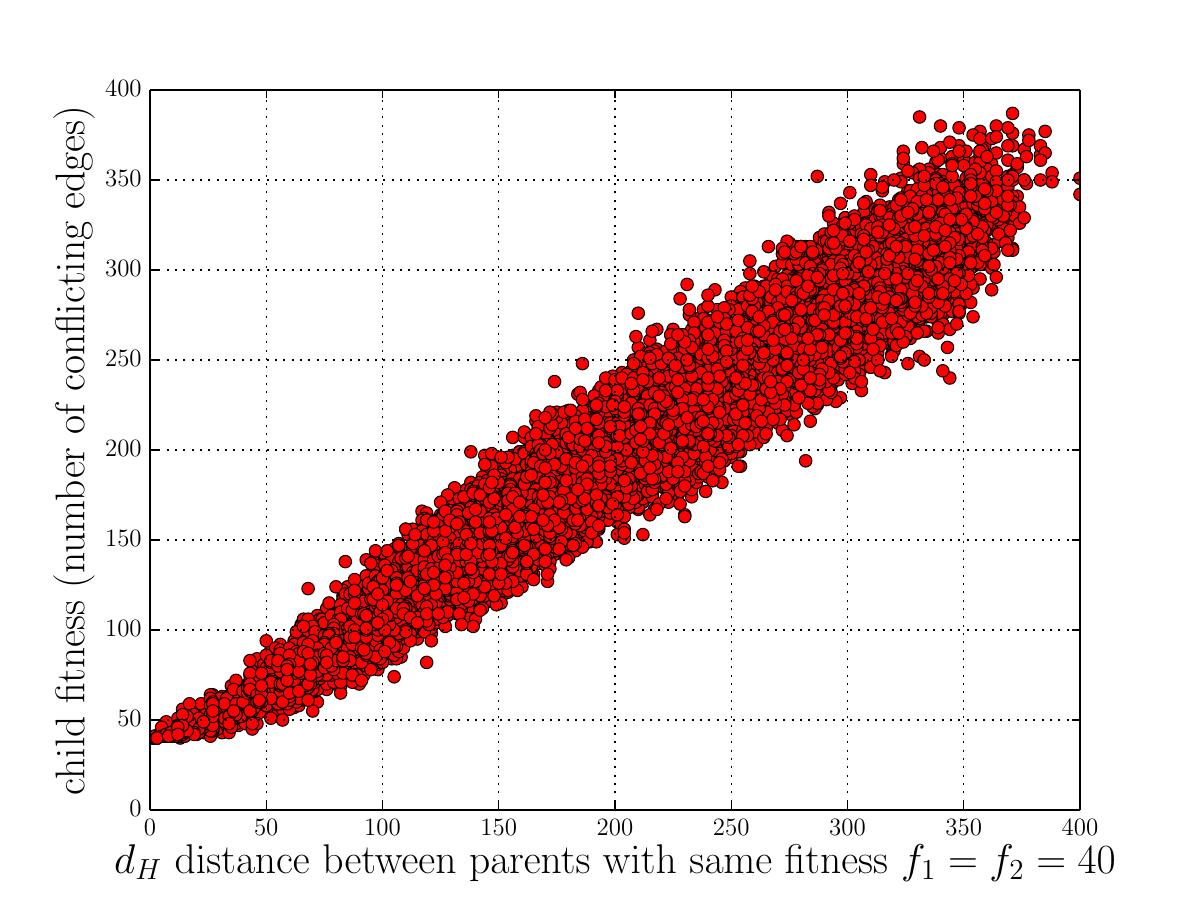}  
 \caption{\label{fig:diversity_crossover}
Each point of the chart, called child-solution, corresponds to one non-legal $48$-coloring of the <\texttt{dsjc500.5}> graph; It has been produced by the \emph{GPX} crossover of two other 
non-legal $48$-colorings, called parents-solutions. Both parents are randomly generated with  
the same fitness value (the same number of conflicting edges). 
Their distance $d_H$ in the search-space is indicated on the abscissa axis.
The number of conflicting edges of the child is indicated on the ordinate axis.
}
\end{center}
 \end{figure}
Moreover, chart~\ref{fig:diversity_crossover} shows that a crossover never improves a $k$-coloring. \supp{: \emph{GPX} is a diversification operator} As stated in section~\ref{sect:GPX}, the last step of \emph{GPX} produces many conflicts.  Indeed, if the two parents are very far in terms of $d_H$, then a large number of vertices remain uncolored at the final step of \emph{GPX}. Those vertices are then randomly added to the color classes, producing many conflicting edges in the offspring.
This explains why in MA, a local search always follows a crossover operator. 

Figure~\ref{fig:running_profile} presents the evolution of the objective function (i.e. the number of conflicting edges) of the two $k$-colorings of the population 
at each generation of \emph{HEAD}. 
It also indicates the $d_H$ distance between the two $k$-colorings.
This figure is obtained by considering one typical run to find one $48$-coloring of <\texttt{dsjc500.5}> graph. 
The objective function of the two $k$-colorings ($f(p_1)$ and $f(p_2)$) are very close during the whole run: the average of the difference $f(p_1)-f(p_2)$ on the $779$ generations is equal to $-0.11$ with a variance of $2.44$.
Figure~\ref{fig:running_profile} shows that there is a significant correlation between the quality of 
the two $k$-colorings (in terms of fitness values) and the distance $d_H$ between them before the \emph{GPX} crossover: 
the Pearson correlation coefficient is equal to 0.927 (respectively equal to 0.930) between $f(p_1)$ and $d_H(p_1,p_2)$ (resp. between $f(p_2)$ and $d_H(p_1,p_2)$).
Those plots give the main key for understanding why \emph{HEAD} is more effective than \emph{HEA}: 
the linear anti-correlation between the two $k$-colorings with approximately same objective function values $f(p_1)\simeq f(p_2)$ is around equal to $500-$$10\times d_H(p_1,p_2)$. 
The same level of correlation with a population of 10 individuals using \emph{HEA} cannot be obtained
except with sophisticated sharing process.


 \begin{figure}[h]
\begin{center}
 \includegraphics[width=.99\linewidth]{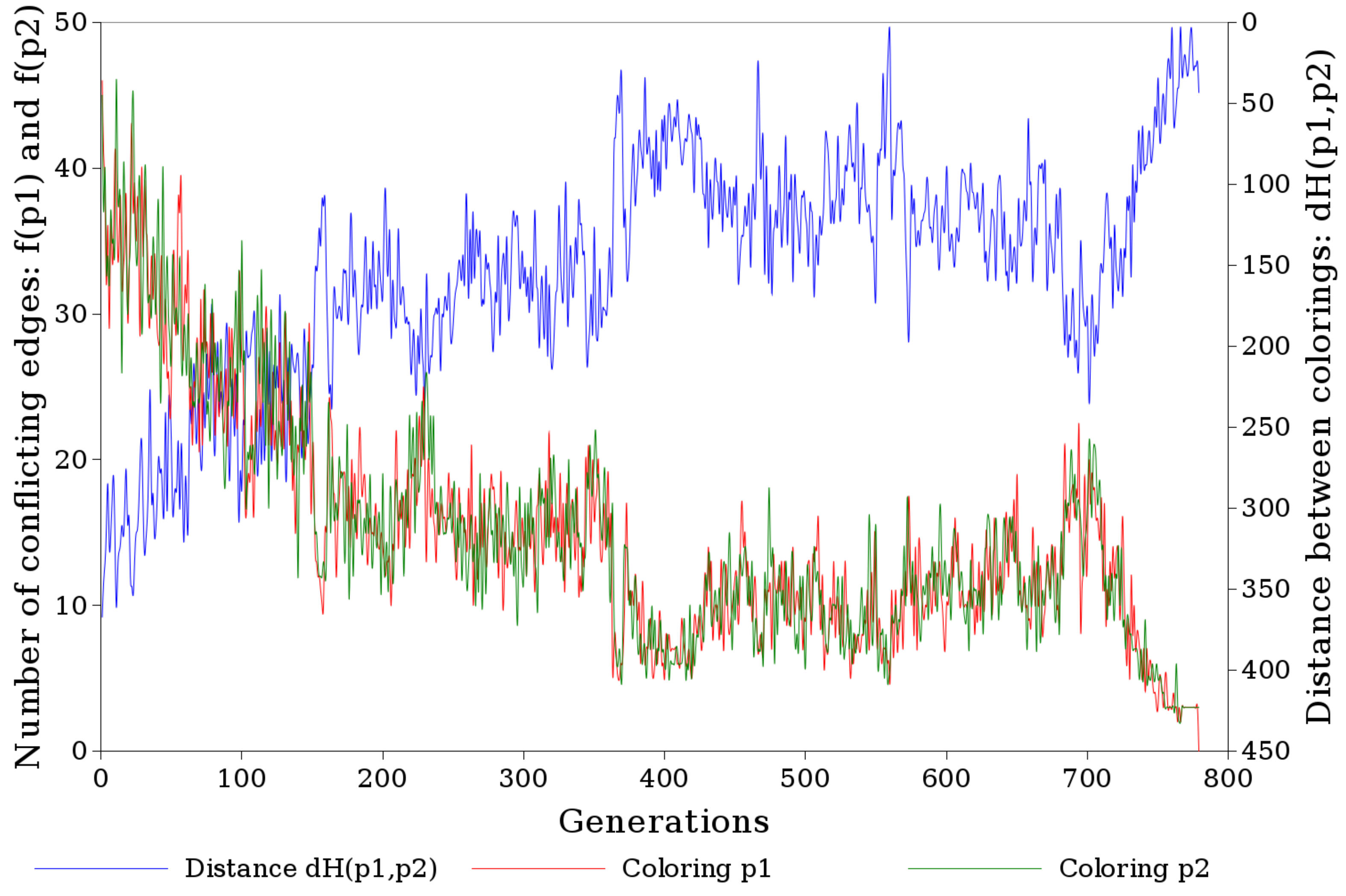}  
 \caption{\label{fig:running_profile}
Red and green lines indicate the number of conflicting edges of the two $48$-colorings of the population: $f(p_1)$ and $f(p_2)$ (left ordinate axis) 
at each generation of one run of \emph{HEAD} applied on <dsjc500.5> graph (abscissa axis).
The blue line indicates the distance between the two $k$-colorings during the run: $d_H(p_1,p_2)$ (right ordinate axis).}
\end{center}
 \end{figure}



Diversity is necessary when an algorithm is trapped in a local minimum but diversity should be avoided in other case.
The next subsections analyze several levers which may able to increase or decrease the diversity in \emph{HEAD}.

\subsection{Dose of diversification in the GPX crossover}\label{subsect:GPX_modification}
Some modifications are performed on the \emph{GPX} crossover in order to increase (as for the first test) 
or decrease (as for the second test) the dose of diversification within this operator.

\subsubsection{Test on GPX with increased randomness: random draw of a number of color classes}\label{subsect:GPX_random}
In order to increase the level of randomness within the \emph{GPX} crossover, we randomize the \emph{GPX}.
It should be remembered (cf. section~\ref{sect:GPX}) that at each step of the \emph{GPX}, the selected parent transmits the largest color class to the child.
In this test, we begin by randomly transmitting $x\in\iintervalleff{0}{k}$ color classes chosen from the parents to the child;
after those $x$ steps, we start again by alternately transmitting the largest color class from each parent 
($x$ is the random level). If $x=0$, then the crossover is the same as the initial \emph{GPX}. 
If $x$ increases, then the randomness and the diversity also increase.
To evaluate this modification of the crossover, we count the cumulative iterations number of \emph{TabuCol}  
that one \emph{HEAD} run requires in order to find a legal $k$-coloring.
For each $x$ value, the algorithm is run ten times in order to produce more robust results.
For the test, we consider the $48$-coloring problem for graph <\texttt{dsjc500.5}> of the DIMACS benchmark.
Figure~\ref{fig:randColor} shows in abscissa the random level $x$ 
and in ordinate the average number of iterations required to find a legal $48$-coloring.

\begin{figure}[h]
\begin{center}
 \includegraphics[width=.5\textwidth]{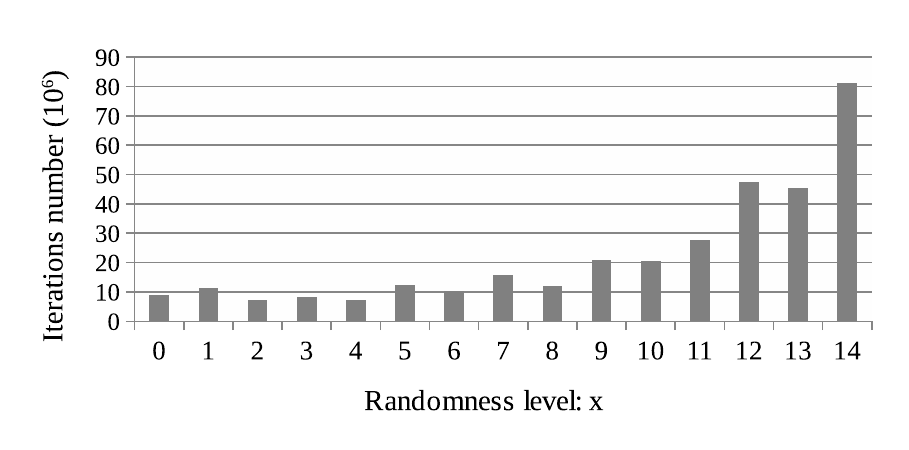}
\caption{\label{fig:randColor}
Average iteration number required to find a legal $48$-coloring for the <\texttt{dsjc500.5}> graph in function of the randomness level;
abscissa: $x$, the randomness level; ordinate: the average iteration number}
\end{center}
\end{figure}

First, $0\leq x\leq k$, where $k$ is the number of colors, but we stop the computation for $x\geq 15$, 
because from $x=15$, the algorithm does not find a $48$-coloring within an acceptable computing time limit.
This means that when we introduce too much diversification, the algorithm cannot find a legal solution.
Indeed, for a high $x$ value, the crossover does not transmit the good features of the parents, therefore the child appears to be a random initial solution. 
When $0\leq x\leq8$, the algorithm finds a legal coloring in more or less 10 million iterations.
It is not easy to decide which $x$-value obtains the quickest result.
However this parameter enables an increase of diversity in \emph{HEAD}.
This version of \emph{GPX} is called \emph{random GPX} and noted R($x$) with $x\in\iintervalleff{0}{k}$ in tables~\ref{tab:resSimpleH2col} and \ref{tab:resH2col}.
It is used for three graphs <\texttt{r250.5}>, <\texttt{r1000.1c}> and <\texttt{dsjr500.1c}> because the standard \emph{GPX} does not operate effectively.
The fact that these three graphs are more structured that the others may explain why the random \emph{GPX} works better.
\subsubsection{Test on GPX with decreased randomness: imbalanced crossover}\label{subsect:GPX_balence}
In the standard \emph{GPX}, the role of each parent is balanced: they alternatively transmit their largest color class to the child.
Of course, the parent which first transmits its largest class, has more importance than the other; this is why it is an asymmetric crossover.
In this test, we give a higher importance to one of the parents. 
At each step of the crossover, we randomly draw the parent that transmits its largest color class with a different probability for each parent.
We introduce $x\in\intervalleff{0}{1}$, the probability of selecting the first parent; $1-x$ is the probability of selecting the second parent.
For example, if $x=0.75$, then, at each step of \emph{GPX}, parent 1 has a 3 in 4 chance of being selected to transmit its largest color class (parent 2 has a 1 in 4 chance).
If $x=0.5$, it means that both parents have an equal probability (a fifty-fifty chance to be chosen); this almost corresponds to the standard \emph{GPX}.
If $x=1$, it means that the child is a copy of parent 1; there are no more crossovers and therefore \emph{HEAD} is a \emph{TabuCol} with two initial solutions.
When $x$ becomes further from the value $0.5$, the chance and diversity brought by the crossover decrease. 
Figure~\ref{fig:imbalancedCrossover} shows in abscissa the probability $x$ and 
in ordinate the average number of necessary iterations required to find a legal $48$-coloring (as in the previous test).

\begin{figure}[h]
\begin{center}
\includegraphics[width=.5\textwidth]{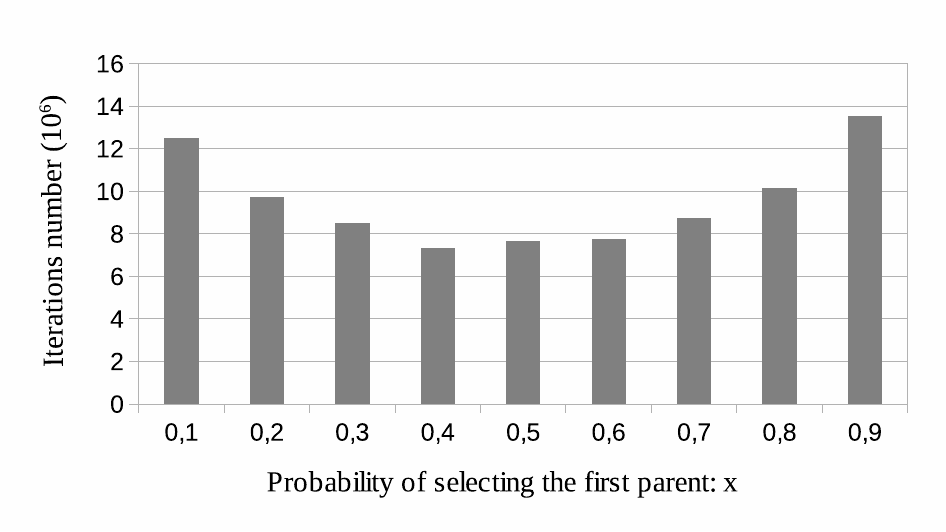}
\caption{\label{fig:imbalancedCrossover}
Average  number of iterations required to find a legal $48$-coloring for <\texttt{dsjc500.5}> graph according to the imbalanced crossover;
abscissa: $x$, probability to select the first parent at each step of \emph{GPX}; ordinate: average iteration number}
\end{center}
\end{figure}

It can be remarked initially that the results are clearly symmetrical with respect to $x$. 
The best results are obtained for $0.4\leq x\leq0.6$.
The impact of this parameter is weaker than that of the previous one: 
the control of the reduction in diversification is finer.
This version of \emph{GPX} is called \emph{unbalanced GPX} and noted U($x$) with $x\in\intervalleff{0}{1}$ in tables~\ref{tab:resSimpleH2col} and \ref{tab:resH2col}.
It is used for three graphs <\texttt{le450\_25c}>, <\texttt{le450\_25d}> and <\texttt{r1000.5}> since the standard \emph{GPX} does not operate effectively.

\subsection{Test on parent replacement: systematic or not}\label{subsect:parent_replacement}
In \emph{HEAD}, the two children systematically replace both parents, even if they have worse fitness values than their parents. 
This replacement rule is modified in this test. 
If the fitness value of the child is lower than that of its parents, the child automatically replaces one of the parents.
Otherwise, we introduce a probability $x$ corresponding to the probability of the parent replacement, even if the child is worse than his parents.
If $x=1$, the replacement is systematic as in standard \emph{HEAD} and
if $x=0$, the replacement is performed only if the children are better (lower fitness value).
When the $x$-value decreases, the diversity also decreases.
Figure~\ref{fig:tauxAcceptWorst} shows in abscissa the parent replacement probability $x$ and 
in ordinate the average number of iterations required to find a legal $48$-coloring (as in the previous test).
\begin{figure}[h]
\begin{center}
\includegraphics[width=.5\textwidth]{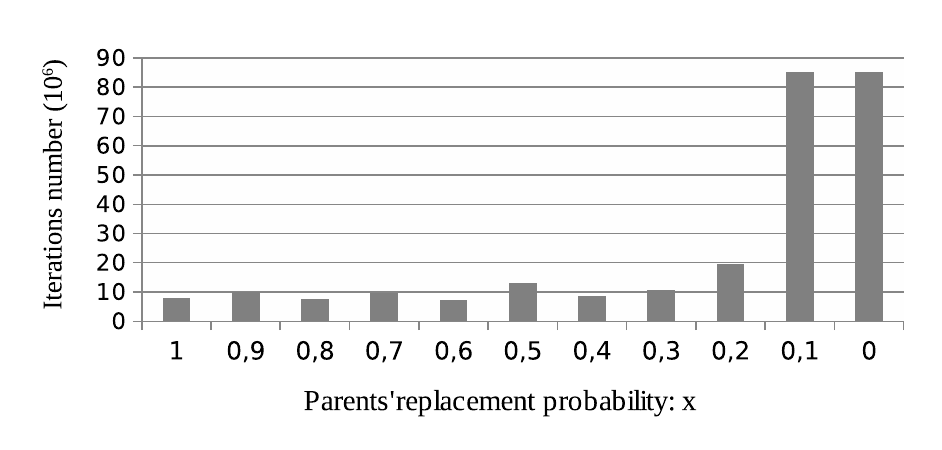}
\caption{\label{fig:tauxAcceptWorst}
Average number of iterations required to find a legal $48$-coloring for <\texttt{dsjc500.5}> graph in function of the parents' replacement policy;
abscissa: parent replacement probability; ordinate: average number of iterations}
\end{center}
\end{figure}
If the parent replacement probability $x=0$ or a very low $0.1$, 
then more time is required to produce the results. 
The absence or the lack of diversification is shown to penalize the search.
However, for a large range of values: $0.3 \leq x \leq 1$, it is not possible to define the best policy for $x$ criterion.
The dramatic change in behavior of \emph{HEAD} occurs very quickly around $0.2$.

These studies enable a clearer understanding of the role of the diversification operators (crossover and parent updating).

The criteria presented here, such as the random level of the crossover or the imbalanced level of the crossover, have shown their efficiency on some graphs. These $GPX$ modifications could successfully be applied into future algorithms in order to manage the diversity dynamically.

\section{Conclusion}\label{sect:conclusion}

We proposed a new algorithm for the graph coloring problem, called \emph{HEAD}. 
This memetic algorithm combines the local search algorithm \emph{TabuCol} 
as an intensification operator with the crossover operator \emph{GPX} as a way to escape from local minima.
Its originality is that it works with a simple population of only two individuals.
In order to prevent premature convergence, the proposed approach introduces an innovative way for managing the diversification based on elite solutions. 


The computational experiments, carried out on a set of challenging DIMACS graphs, show that \emph{HEAD} produces accurate results, such as 222-colorings for <\texttt{dsjc1000.9}>, 81-colorings for <\texttt{flat1000\_76\_0}> and even 47-colorings for <\texttt{dsjc500.5}> and 82-colorings for <\texttt{dsjc1000.5}>, which have up to this point only been found by quantum annealing~\cite{Titiloye12} with a massive multi-CPU.
The results achieved by \emph{HEAD} let us think that this scheme could be successfully applied to other problems, where a stochastic or asymmetric crossover can be defined.


We performed an in-depth analysis on the crossover operator in order to better understand its role in the diversification process.
Some interesting criteria have been identified, such as the crossover's levels of randomness and imbalance.
Those criteria pave the way for further researches.

\bibliographystyle{elsarticle-num}
\bibliography{gcp_bib.bib}

\end{document}